\documentclass[fleqn,10pt,twocolumn]{AROB}
\usepackage[T1]{fontenc}
\usepackage{newtxtext}

\usepackage{amsmath} 
\usepackage{amsfonts}
\usepackage[normalem]{ulem}

\usepackage{tabularx}
\usepackage{algorithm}
\usepackage{algorithmic}

\usepackage{makecell}
\usepackage{tcolorbox}
\usepackage{cite}

\tcbuselibrary{skins, breakable, theorems}
\usepackage{cleveref}
\newcommand{\kara}{}
\newtcolorbox[auto counter, number within=section, crefname = {Prompt.}{Prompts.}]{prompt}[3][]
{enhanced, breakable = true, fonttitle = \bfseries, fontupper=\scriptsize,   fontlower = \scriptsize, left=2mm, right=2mm, top=1mm, bottom=1mm, float, floatplacement=tb,
title = Prompt.~\thetcbcounter.~\if #2\kara \else #2 \fi,
#1,
label = prompt:#3}

\title{Toward Ownership Understanding of Objects: Active Question Generation \\
with Large Language Model and Probabilistic Generative Model
}

\author{
Saki Hashimoto${}^{1\dagger}$ ,
Shoichi Hasegawa${}^{1}$,
Tomochika Ishikawa${}^{1}$,
Akira Taniguchi${}^{2}$, \\%
Yoshinobu Hagiwara${}^{3,4}$, %
Lotfi El Hafi${}^{4}$, %
and Tadahiro Taniguchi${}^{4,5}$%
}

\speaker{Saki Hashimoto}

\affils{
${}^{1}$Graduate School of Information Science and Engineering, Ritsumeikan University, Osaka, Japan\\
(E-mail: \{ hashimoto.saki, hasegawa.shoichi, ishikawa.tomochika \} @em.ci.ritsumei.ac.jp)\\
${}^{2}$College of Information Science and Engineering, Ritsumeikan University, Osaka, Japan\\
(E-mail: a.taniguchi@em.ci.ritsumei.ac.jp)\\
${}^{3}$Faculty of Science and Engineering, Soka University, Tokyo, Japan\\
(E-mail: hagiwara@soka.ac.jp)\\
${}^{4}$Research Organization of Science and Technology, Ritsumeikan University, Shiga, Japan\\
(E-mail: lotfi.elhafi@em.ci.ritsumei.ac.jp)\\
${}^{5}$Graduate School of Informatics, Kyoto University, Kyoto, Japan\\
(E-mail: taniguchi@i.kyoto-u.ac.jp)\\
}

\abstract{
Robots operating in domestic and office environments must understand object ownership to correctly execute instructions such as ``Bring me my cup.'' 
However, ownership cannot be reliably inferred from visual features alone. 
To address this gap, we propose Active Ownership Learning~(ActOwL), a framework that enables robots to actively generate and ask ownership-related questions to users.
ActOwL employs a probabilistic generative model to select questions that maximize information gain, thereby acquiring ownership knowledge efficiently to improve learning efficiency.
Additionally, by leveraging commonsense knowledge from Large Language Models~(LLM), objects are pre-classified as either shared or owned, and only owned objects are targeted for questioning. 
Through experiments in a simulated home environment and a real-world laboratory setting, ActOwL achieved significantly higher ownership clustering accuracy with fewer questions than baseline methods. 
These findings demonstrate the effectiveness of combining active inference with LLM-guided commonsense reasoning, advancing the capability of robots to acquire ownership knowledge for practical and socially appropriate task execution.
}


\begin{document}

\maketitle


\section{Introduction}\label{introduction}
Robots operating in daily life environments must understand object ownership to carry out instructions naturally given by users, such as ``Bring me my cup.'' 
Without ownership knowledge, a robot cannot determine which object is being referred to when multiple similar objects exist. 
This problem is especially evident in kitchens, offices, or laboratories, where objects with similar appearances may belong to different individuals. 
Relying solely on perceptual features such as location or appearance is insufficient because ownership is inherently context-dependent and often determined by social conventions. 
Therefore, enabling robots to acquire ownership knowledge is a crucial step toward socially appropriate human-robot interaction.

To enable robots to learn object ownership in daily life environments, it is essential to implement a question-generation mechanism that efficiently acquires necessary information.
However, in real-world environments with large numbers of objects, this is impractical and imposes a heavy burden on users.
Although robots can explore the environment to collect visual features of objects, it remains difficult to obtain ownership knowledge because it depends on users and context.
Therefore, allowing robots to ask questions based on the current situation enables them to acquire ownership knowledge through interaction with users.

\begin{figure}[tb]
    \centering
    \includegraphics[width=1.0\linewidth]{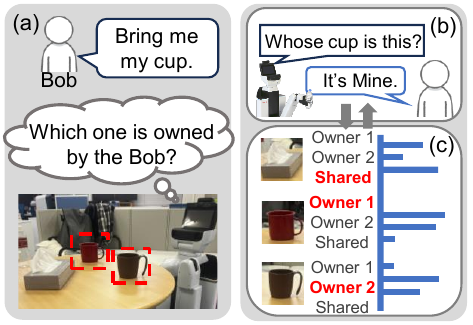}
    \caption{
    Overview of this study:
    (a) Without ownership knowledge, the robot cannot follow instructions containing owner names.
    (b) The robot generates questions for the user, and
    (c) The probabilistic generative model is updated to predict object ownership based on the user's answers, enabling ownership learning
    }
    \label{research_image}
\end{figure}

Efficiency can be improved by first distinguishing shared from owned objects. 
In real-world environments, many objects, such as tissue boxes, are typically shared rather than individually owned. 
Asking about such objects increases cognitive load and lowers learning efficiency. 
By pre-classifying objects as shared or owned and excluding shared objects from questioning, the robot can minimize interactions and acquire ownership knowledge more effectively.

In learning ownership knowledge, it is important for the robot to integrate perceptual features—such as object location and visual attributes—with ownership labels provided by users, and to store this information as structured, linked knowledge.
In this study, we hypothesize that objects owned by the same person tend to be placed in close proximity and share similar attributes.
Based on this hypothesis, we assume that the robot can achieve more accurate ownership estimation by associating spatial and visual features with owner identities.
For example, by associating spatial and visual characteristics (e.g., ``a small, round, brown cup located at coordinates (1, 2)'') with the corresponding owner (e.g., ``owned by Taro''), the robot can accurately learn object ownership relationships.

Our challenge in this study is to improve the efficiency of learning object ownership knowledge in daily life environments through robot-generated questions.
To overcome this challenge, this study introduces Active Ownership Learning~(ActOwL), a framework that enables robots to efficiently learn ownership knowledge through active question generation. 
An overview of this study is shown in Fig.~\ref{proposed_method_image}.
ActOwL integrates a probabilistic generative model with commonsense reasoning from Large Language Models (LLM). 
First, objects are pre-classified as shared or owned based on LLM-guided commonsense knowledge, allowing the robot to avoid asking questions about shared objects. 
Then, for candidate-owned objects, ActOwL strategically selects the most informative question by maximizing information gain~(IG) and generates human-like questions via LLM. 
By incorporating the principle of active exploration into the question generation process and selecting objects that maximize IG~\cite{EFE,SpCoAE,ActiveSpCoSLAM}, ActOwL effectively reduces model uncertainty during ownership learning. 
The user's answers are subsequently incorporated into the probabilistic generative model, enabling the robot to incrementally refine its ownership knowledge. 

To evaluate the effectiveness of the ActOwL, we conducted experiments in simulated and real-world environments that mimic home and laboratory settings. 
In addition, we quantitatively compared ownership clustering accuracy and the number of questions required, and further performed ablation studies on the probabilistic generative model to analyze the impact of multimodal attributes on learning efficiency.

The main contributions of this study are as follows:
\begin{enumerate}
\item 
We show a probabilistic generative model that integrates object location, attribute information, and user answers to learn ownership distributions, demonstrating that combining these multimodal inputs is effective for acquiring ownership knowledge.
\item 
We show an active object selection method based on IG, and clarify how the selection strategy influences the efficiency of ownership learning.
\item 
We show through comparative experiments with baseline and ablation methods in both simulation and real-world environments that the ActOwL achieves higher ownership clustering accuracy with fewer questions.
\end{enumerate}

The structure of this paper is as follows.
Section~\ref{problem_statement} clarifies the problem setting and key challenges of this study.
Section~\ref{related_work} reviews related work.
Section~\ref{proposed_method} describes the details of the ActOwL.
Sections~\ref{exp1}, \ref{exp2}, and~\ref{exp3} present experimental results that evaluate ownership learning from the perspectives of clustering accuracy, quantified by ARI, and learning efficiency, assessed by the number of questions required.
Section~\ref{limitaion} presents the limitations of this study.
Finally, Section~\ref{conclusion} concludes the paper and discusses future directions.

\section{Problem Statement}\label{problem_statement}
This study focuses on the following two key approaches:
\begin{enumerate}
\item 
Improving ownership learning by constructing a probabilistic generative model that integrates object location, attributes, and user answers.
\item 
Enhancing ownership learning efficiency by actively selecting objects for questioning based on IG.
\end{enumerate}

\subsection{Challenges in Handling ownership knowledge}
For robots to behave appropriately in daily life environments, they must accurately understand object ownership. 
However, learning ownership knowledge based solely on spatial location or visual attributes has inherent limitations.
For instance, in real-world settings, even objects owned by the same person may be placed in different locations due to their intended use or storage convenience, leading to inconsistencies in spatial placement.
As such, observable information such as location and appearance alone is often insufficient to uniquely identify the owner of an object.
Therefore, in addition to exploring the environment, robots must engage in interaction with users to obtain supplementary information, thereby accurately acquiring ownership knowledge.

\subsection{Challenges in Acquiring Information through Questioning}
To learn object ownership in daily life environments, it is effective for a robot to obtain necessary information through interaction with users.
However, asking users to manually provide ownership knowledge for every object in the environment imposes a significant burden and is impractical in real-world settings.
Moreover, if the robot generates questions randomly, the efficiency of learning ownership knowledge may be substantially reduced.
Therefore, a strategic approach is required, in which the robot selectively asks questions about objects that are likely to yield informative answers, while minimizing the number of interactions.

\section{Related Work}\label{related_work}
\subsection{Learning Object Ownership}
Understanding object ownership is essential for assistive robots, and several approaches have been proposed to address this challenge. Prior work has estimated ownership from object attributes, spatial features, or user interactions~\cite{tan2019s,wu2020item,hu2023interactive}. While these methods provide useful cues, they typically lack integration of multimodal information and do not address efficient question generation, commonsense reasoning, or autonomous decision-making in real-world environments.

In contrast, our approach enables robots to autonomously acquire ownership knowledge through interaction with users. By leveraging commonsense knowledge from LLM and a probabilistic generative model that integrates object location, attributes, and user answers, our method selects informative questions efficiently. This allows robots to minimize user burden while acquiring ownership knowledge directly applicable to task execution in daily environments.

\subsection{Knowledge Acquisition through Active Exploration}
Beyond passive observation, robots must actively explore their environments to reduce uncertainty and acquire useful knowledge under real-world constraints. 
Prior studies have introduced IG-based exploration strategies, including multimodal perception with the Multimodal Hierarchical Dirichlet Process~(MHDP)~\cite{taniguchi2018multimodal,MHDP} and spatial concept learning frameworks~\cite{SpCoAE,ActiveSpCoSLAM}. 
These approaches integrate multimodal information and exploit IG to guide efficient exploration, yet they primarily aim to understand the environment from observable cues such as appearance, sound, and location. 
However, they do not consider ownership relations between users and objects, which are crucial for executing user-specific commands.

In contrast, our study incorporates the uncertainty reduction principle of active exploration into the question generation process, applying IG-based selection to identify the most informative objects. 
By combining this with a probabilistic generative model that integrates object position, attributes, and user answers, our approach enables robots to minimize user burden while acquiring ownership knowledge directly applicable to task execution.

\subsection{Resolving Uncertainty during Task Execution}
Recent work has explored methods for resolving uncertainty in task execution from natural language instructions. 
Systems prompt clarification questions to disambiguate commands~\cite{KNOWNO,Clara}, while other approaches combine perceptual cues, contextual reasoning, or language models to identify targets under uncertainty~\cite{GoFind,TtR,ORION,OpenFMNav,MIEL}. 
Another line of work further incorporates user preferences through active questioning~\cite{apricot}.
These methods demonstrate that robots can flexibly address ambiguity through interaction and inference, enabling robust task execution in complex environments.

However, ownership relations are typically treated as temporary and task-specific, without being structured or retained as reusable knowledge. As a result, repeated instructions require the re-acquisition of the same information, and sequential questioning often increases the user's burden. In contrast, our study enables robots to autonomously acquire and retain structured ownership knowledge by integrating object location, attributes, and user answers in a probabilistic generative model. This allows efficient interpretation of ambiguous instructions while minimizing redundant user interactions.

\section{Proposed Method}\label{proposed_method}
In this study, we propose Active Ownership Learning~(ActOwL), a method that enables robots to efficiently learn object ownership in daily life environments. 
ActOwL selects objects with the highest IG and asks targeted ownership questions, reducing the need for extensive manual instruction. 
Ownership knowledge is then acquired at the category level through a probabilistic generative model that integrates multimodal information, including object location, attributes, and user answers.

\begin{figure*}[tb]
    \centering
    \includegraphics[width=1.0\linewidth]{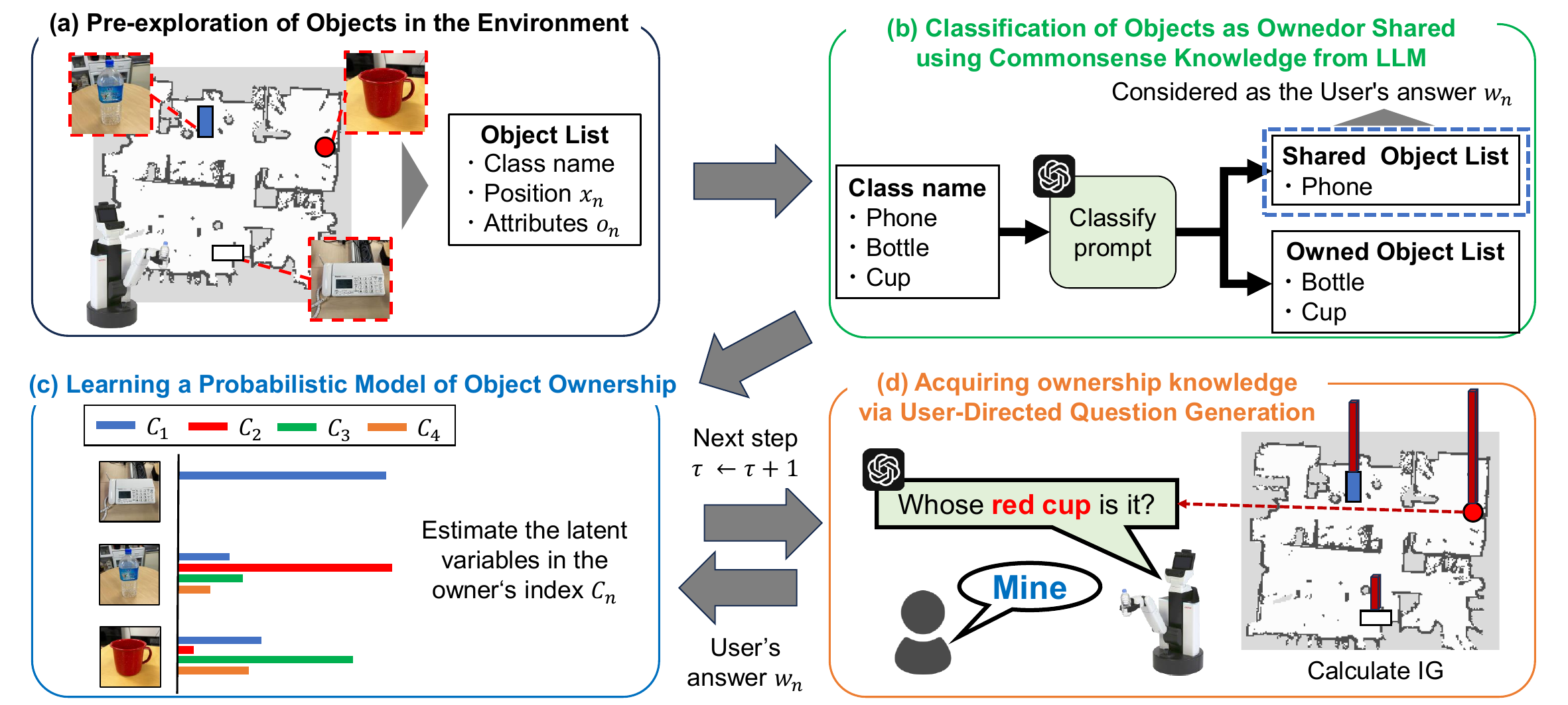}
    \caption{
    Overview of the proposed method:
    (a) The robot explores the environment to obtain object positions and attributes, training an initial ownership model.
    (b) Objects are classified as owned or shared using commonsense knowledge from the LLM.
    (c) Classification results are treated as pseudo-answers to update the ownership distribution.
    (d) For owned objects, the robot computes IG, selects the most informative one, generates a question via LLM, and updates the model based on the user's answer.
    }
    \label{proposed_method_image}
\end{figure*}

An overview of the proposed method is shown in Fig.~\ref{proposed_method_image}.
The procedure consists of the following steps:
\begin{enumerate}
\item 
Train a probabilistic generative model of ownership using the position and attribute information of each object obtained through prior exploration (Fig.~\ref{proposed_method_image}(a)).
\item 
Classify each object as owned or shared based on commonsense knowledge from the LLM (Fig.~\ref{proposed_method_image}(b)).
\item 
Treat the classification results as pseudo-user answers and use them to update the generative model (Fig.~\ref{proposed_method_image}(c)).
\item 
For objects predicted as owned, compute the IG regarding ownership and identify the object expected to reduce uncertainty most effectively (Fig.~\ref{proposed_method_image}(d)).
\item 
Generate a natural question about the selected object using the LLM and ask the user (Fig.~\ref{proposed_method_image}(d)).
\item Update the generative model with the user's answer (Fig.~\ref{proposed_method_image}(c)).
\item 
Recalculate IG based on the updated model and select the next object for questioning (Fig.~\ref{proposed_method_image}(d)).
\end{enumerate}
These steps are repeated until the ownership of all objects is identified.
Details of the prior environmental exploration are described in Section~\ref{ple_exploration},
the classification of objects as shared or owned in Section~\ref{classify_object},
the learning of ownership distributions in Section~\ref{model_learning},
and the object selection and question generation process in Section~\ref{sec:question_generation}.

\subsection{Pre-exploration of Objects in the Environment}\label{ple_exploration}
In this study, we assume that the robot has prior knowledge of the spatial layout and categories of objects in the environment to efficiently learn ownership knowledge. 
The robot is provided with an environment map and object-related data, including class names, 2D coordinates, visual attributes, and a list of candidate owners. 
Object positions are represented as coordinates on the robot's 2D map, and visual attributes are encoded as feature vectors that represent the appearance. 
Such representations can be obtained using vision–language models, such as CLIP~\cite{CLIP}, while spatial and attribute information can be embedded through semantic mapping techniques, such as NLMap~\cite{NLMap}. 
The candidate owner list specifies potential owners with concrete user names. Using this information, the robot strategically generates questions based on spatial configuration and inter-object relationships to efficiently acquire ownership knowledge.

\subsection{Classification of Objects as Owned or Shared using Commonsense Knowledge from LLM}\label{classify_object}

\begin{prompt}{Object Ownership Classification}{classify_object_prompt}
1: You are an excellent household robot.\\
2: Please classify the list of objects that we give you as either ``owned'' or ``shared'' by someone in the living environment.\\
3: Let's think step by step.\\
4:\\
5: A list of the objects you observed in a home environment:\\
6: existing\_object = [\textcolor{teal}{\textbf{OBJECT\_LIST}}]\\
7:\\
8: However, the environment is a family household.\\
9:\\
10: Output only those objects that are property as a result of classification.\\
11:\\
12: Follow the instructions below to answer the questions.\\
13: Do not output anything other than the answer.\\
14:\\
15: Example: If the list of objects is\\
16: [book, pen, desk, window].\\
17:\\
18: Answer:\\
19: Owned\_object = [book, pen]
\tcblower 
\textcolor{teal}{\textbf{OTHER\_OBJECT\_LIST}}: List of Objects in the Environment\\ 
(Class, Attribute, Point.X, Point.Y, Observed User)
\end{prompt}

To improve the efficiency of ownership learning, we introduce a filtering mechanism that pre-selects objects for questioning. 
Objects are classified as either owned or shared using commonsense knowledge from an LLM. 
As illustrated in Prompt.~\ref{prompt:classify_object_prompt}, the robot provides the object's class name as input to the LLM, which determines whether the object is typically shared or individually owned. 
Objects classified as shared are assumed not to have a specific owner and are excluded from question generation and probabilistic modeling, whereas objects classified as owned are retained as candidates for ownership identification.

\subsection{Learning a probabilistic generative model of Object Ownership}\label{model_learning}
\begin{figure}[tb]
    \centering
    \includegraphics[width=1.0\linewidth]{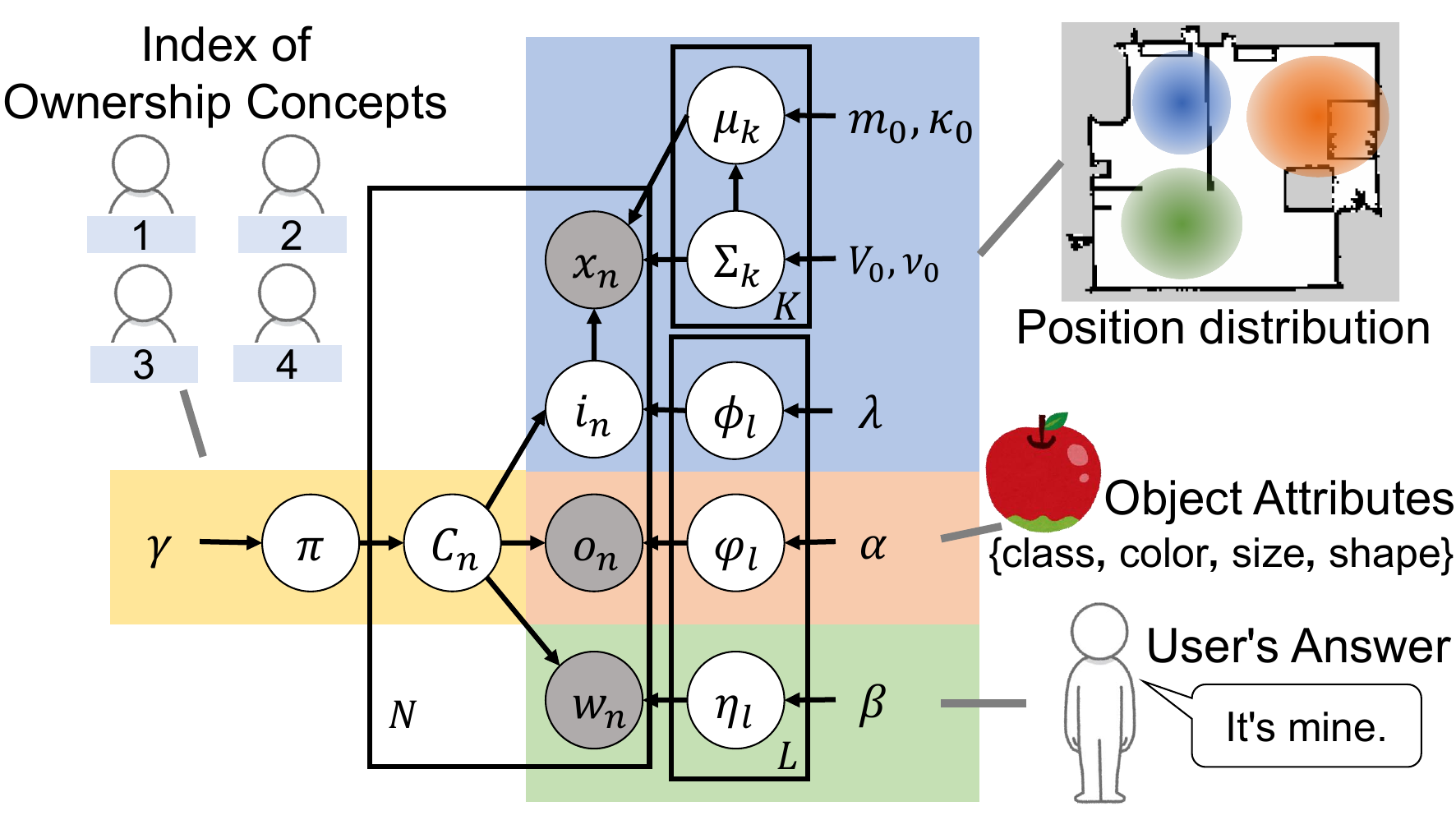}
    \caption{Graphical model of the proposed method}
    \label{model}
\end{figure}

The robot learns a probabilistic generative model to estimate object ownership from the position and attribute information obtained during prior exploration. 
An overview of the model is illustrated in Fig.~\ref{model}, and the variables are defined in Table~\ref{variable definition}. 
The generative process is formalized as follows.

\begin{align}
    \pi         &\sim {\mathrm{Dir}}(\gamma) &&  \\
    \Sigma_k    &\sim {\mathcal{IW}}(V_0,\nu_0) && k=1,2,\dots, K \\
    \mu_k       &\sim {\mathcal{N}}(m_{0},\Sigma_k /\kappa_0) && \\
    \phi_l      &\sim {\mathrm{DP}}(\lambda) && l=1,2,\dots, L \\    
    \varphi_l   &\sim {\mathrm{Dir}}(\alpha) && \\
    \eta_l      &\sim {\mathrm{Dir}}(\beta) && \\
    C_n         &\sim {\mathrm{Cat}}(\pi) && n=1,2,\dots, N \\
    i_n         &\sim {\mathrm{Cat}}(\phi_{C_n}) && \\
    x_n         &\sim {\mathcal{N}} (\mu_{i_n}, \Sigma_{i_n}) && \\
    o_n         &\sim {\mathrm{Mult}}(\varphi_{C_n}) && \\
    w_n         &\sim {\mathrm{Mult}}(\eta_{C_n}) &&
\end{align}

$\mathrm{Dir}()$ denotes the Dirichlet distribution, $\mathrm{DP}()$ the Dirichlet process, $\mathrm{Mult}()$ the multinomial distribution, and $\mathrm{Cat}()$ the categorical distribution.
$\mathcal{IW}()$ denotes the inverse-Wishart distribution, and $\mathcal{N}()$ denotes the multivariate Gaussian distribution.

The position distribution assumes that a person's belongings are more likely to be found in their own room. To handle shared spaces such as kitchens and living rooms used by multiple individuals, it is modeled as a Gaussian mixture. The attribute vector includes not only the object's class but also features such as color, size, and shape, allowing the model to capture individual user preferences. For example, if a user typically prefers red objects, red objects are more likely to be inferred as belonging to that person.

User answers obtained through question generation do not always correspond directly to predefined owner indices. For instance, instead of explicitly naming an owner (e.g., ``Taro's''), users may reply with pronouns such as ``mine'' or expressions like ``my father's.'' To address this variability, the system maps user responses to ownership indices through semantic interpretation.

\begin{table}[b]
\centering
\caption{Definitions of variables used in the graphical model of the proposed method}
\label{variable definition}
\begin{tabularx}{1.0\linewidth}{|c|X|}
\hline
$C_n$ & Latent variable indicating the ownership index \\ \hline
$x_n$ & 2D coordinates of the object \\ \hline
$o_n$ & Attribute vector of the object \\ \hline
$w_n$ & User answer to the robot's question \\ \hline
$i_n$ & Latent variable indicating the index of the position distribution \\ \hline
$\pi, \varphi_l, \eta_l, \phi_l$ & Parameters of multinomial distributions \\ \hline
$\Sigma_k, \mu_k$ & Mean vector and covariance matrix of the position distribution \\ \hline
\shortstack[l]{
$\gamma$, $m_0$, $\kappa_0$, $V_0$, \\
$\nu_0$, $\lambda$, $\alpha$, $\beta$
} & Hyperparameters of the model \\ \hline
$n$ & Index of an object \\ \hline
$N$ & Total number of objects obtained during prior exploration \\ \hline
$l$ & Index of an ownership concept \\ \hline
$L$ & Total number of ownership concepts \\ \hline
$k$ & Index of a position distribution component \\ \hline
$K$ & Total number of position distribution components \\ \hline
\end{tabularx}
\end{table}

To incrementally update the ownership distribution in answer to sequential observations, we adopt an online learning approach using a Rao-Blackwellized Particle Filter~(RBPF)~\cite{Online_learning}, following the framework of SpCoAE~\cite{SpCoAE}.
The robot treats the object's position, attributes, and the user's answer as multimodal observations and performs sequential inference of the posterior distribution over ownership.


The joint posterior distribution of all parameters estimated during ownership learning is factorized using Bayes' theorem.
Following the derivation presented in SpCoAE~\cite{SpCoAE}, the particle weight update equation can be expressed as follows:
\begin{align}
\begin{split}
    & p(x_n, o_n, w_n | C_{1:n-1}, i_{1:n-1} , x_{1:n-1}, o_{1:n-1}, w_{1:n-1},h)\\
    & \quad=\sum_{C_n}\sum_{i_n} p(x_n | x_{1:n-1},i_{1:n},h) p(o_n | o_{1:n-1}, C_{1:n},\alpha) \\
    & \quad \quad \times p(w_n | w_{1:n-1}, C_{1:n},\beta)\\ & \quad \quad \times p(C_{1:n}, i_{1:n} | C_{1:n-1}, i_{1:n-1}, \gamma, \lambda)
\end{split}
\end{align}
Based on this equation, particles are resampled according to their weights $\omega_n$.
During the initial exploration phase, the robot does not ask any questions; therefore, $w_n$ is treated as uninformative.
After question generation begins, the probabilistic generative model is updated by incorporating the user's answer $w_n$, enabling the learning of ownership knowledge.

\subsection{Acquiring ownership knowledge via User-Directed Question Generation}
\label{sec:question_generation}
To efficiently acquire ownership knowledge, the proposed method calculates IG for each object and selects the one expected to most effectively reduce uncertainty.
For the selected object, a natural language question is generated using LLM, and an answer is obtained from the user. 

\subsubsection{Object Selection Based on IG}
\label{select_by_IG}
For objects identified as owned belongings, the next object to query is selected based on its expected IG.
The overall algorithm for the proposed method is presented in Algorithm~\ref{alg_IG}.
Here, \textbf{UPDATE-MODEL} denotes the RBPF-based update step, which takes the current set of observations $(x_n, o_n, w_n)$ and sequentially updates the posterior distribution of ownership concepts and model parameters.


\begin{figure}[!t]
\begin{algorithm}[H]
    \caption{Active Exploration Algorithm for Ownership Acquisition}
    \label{alg_IG}
    \begin{algorithmic}[1]
    \STATE $Z_{-1} = \text{UPDATE-MODEL}\big(\{w_a=\text{unknown} \mid \forall a\}\big)$
    \STATE Identify shared objects via LLM; set $w_a=\text{Shared}$
    \STATE $Z_0 = \text{UPDATE-MODEL}\big(\{w_a \in \{\text{unknown, Shared}\} \mid \forall a\}\big)$
    \STATE $n_0 = \{a \mid w_a=\text{Shared}\}$,\quad $W_{n_0} = \{w_a \mid a \in n_0\}$

    \FOR{$\tau = 1$ to $ \mathcal{T}$}  
        \FORALL{detected objects $x_a, o_a \in \mathbb{R}^D$}  
            \FOR{$r=1$ to $R$}  
                \FOR{$j=1$ to $J$} 
                    \STATE $W_a^{[r,j]} \sim \text{Multinomial} (p=\hat{p}(w_a|Z^{[r]},W_{n_0}))$
                \ENDFOR
            \ENDFOR
            \STATE Compute $\text{IG}_a$ using sampled $W_a^{[r,j]}$ and particle weights $\omega^{[r]}_{n_0}$
        \ENDFOR
        \STATE $a^\star \gets \arg\max_{a} \mathrm{IG}_a$
        \STATE $q_{a^*} = \text{QUESTION-GENERATION}(W_{a^*})$
        \STATE $w_{a^\star} \gets \textsc{UNDERSTAND-ANSWER}(a^\star)$
        \STATE $n_0 \gets n_0 \cup \{a^\star\}$,\quad $W_{n_0} \gets W_{n_0} \cup \{w_{a^\star}\}$
        \STATE $Z_{\tau} \gets \text{UPDATE-MODEL}(W_{n_0})$
    \ENDFOR
    \end{algorithmic}
\end{algorithm}
\end{figure}

In this study, the full set of parameters related to ownership is denoted by $\Theta=\{\{\mu_k\},\{\Sigma_k\},\{\phi_l\},\{\varphi_l\},\{\eta_l\},\pi \}$, and the set of hyperparameters is denoted by $h=\{ \alpha, \beta,\gamma,\lambda, m_0, \kappa_0, V_0, \nu_0 \}$.
The latent variables are represented as $Z=\{C_{1:N},i_{1:N},\Theta\}$, and Set of observed information is $W_{n_0}=\{x_{1:N},o_{1:N},w_{n_0},h\}$.
Here, $a\in \{1:N\} \setminus n_0$ denotes the index of a candidate object for the next question, while $n_0$ represents the set of indices corresponding to objects that have already been observed.

IG is defined to reduce the uncertainty in the ownership distribution before and after a question is asked.
\begin{align}
\begin{split}
    a^*
    &=\underset{a}{\text{argmax}} \text{IG}(Z;W_a|W_{n_0})
\end{split}
\end{align}

IG is approximated using the particles $Z^{[r]}$ and their corresponding weights $\omega^{[r]}_{n_0}$ obtained from the RBPF, as expressed by the following equation:

\begin{align}
    &IG(Z;W_a|W_{n_0}) \nonumber \\
    &\approx \sum^{R}_{r=1} \sum^{J}_{j=1}\left[\omega^{[r]}_{n_0} \log \frac{p(W_a^{[j]}|Z^{[r]},W_{n_0})}{\sum^{R}_{r'=1}[p(W_a^{[j]}|Z^{[r']},W_{n_0})]\omega^{[r']}_{n_0}]}\right]
\end{align}

\begin{align}
    W_a^{[j]} &\sim p(W_a|Z^{[r]},W_{n_0})
\end{align}

$W_a^{[j]}$ denotes a pseudo-observation (i.e., a simulated user answer $w_a$) for object $a$ obtained from the next possible question, and $J$ represents the number of pseudo-observation samples.
The term $p(W_a^{[j]} \mid Z^{[r]}, W_{n_0})$ is sampled from the predictive distribution based on particle $r$, and is computed using the following equation:

\begin{align}
\begin{split}
    W_a^{[j]} 
    &= (w_a |x_a, o_a) \sim p(w_a | Z^{[r]}, x_a, o_a, w_{n_0},a,h)\\
    &= \sum_{C_a,i_a} p(w_a|C_a,w_{n_0},h) p(C_a,i_a | Z^{[r]}, x_a, o_a, a, h)
\end{split}
\end{align}

The position $x_a$ and attribute $o_a$ of the object are held fixed, and sampling is performed only over the user answer $w_a$.
By leveraging the particles and weights obtained through online learning, IG can be efficiently approximated.

\subsubsection{Question Generation using LLM}\label{QG_by_LLM}
For the object with the highest IG, the robot generates a question about ownership. 
Conventional approaches to question generation have relied on predefined templates~\cite{TtR} or on masking portions of object datasets to elicit answers~\cite{K-VQG}. 
However, these methods are often inflexible in novel environments with unknown object configurations, and the resulting questions tend to be overly formal and restrictive, limiting the naturalness and diversity of user responses.

To address this issue, our approach employs an LLM, as in prior studies~\cite{KNOWNO,Clara,ORION}, to generate questions in a more natural and conversational manner. 
This enables flexible question generation and intuitive interaction with users. When generating questions, the LLM receives not only the attributes of the target object but also its coordinates, features, and ownership knowledge already learned for other objects in the environment. 
The prompt used for this process is shown in Prompt.~\ref{prompt:ask_user_prompt}, which allows the model to incorporate spatial relationships and prior knowledge to produce contextually appropriate questions.

\begin{prompt}{Object Ownership Question Generation}{ask_user_prompt}
1: You are an excellent household robot.\\
2: Generate a question asking the owner of the object we provide.\\
3: In addition, provide information about other objects observed in the home environment.\\
4: Note that similar objects may have been placed nearby, or objects near one with a known owner may share the same owner.\\
5: Instead of simply asking ``Whose is this?'', add brief object information to help the listener understand the question.\\
6: Let's consider them in this order.\\
7:\\
8: Below is information about each object. Use the following format to describe it:\\
9: [class, attribute, point.x, point.y, observed user]\\
10: The description of each variable is as follows\\
11: class: Class name of the object\\
12: attribute: Feature vector of an object\\
13:  The vector concatenates: Object label (18D) +\\
14:  Attribute [color: red, blue, yellow, green, black, white] (6D) +\\
15:  Attribute [size: large, medium, small] (3D) +\\
16:  Attribute [shape: round, square, triangle] (3D).\\
17: The object labels are:\\
18: [Backpack, Bed, Book, Bottle, Chair, Clock, Cup, Desk, Dining Table, Handbag/Satchel, Laptop, Monitor/TV, Mouse, Pillow, Potted Plant, Printer, Refrigerator, Trash bin Can].\\
19: answer: owner name of the object\\
20: point.x: x-coordinate of object\\
21: point.y: y-coordinate of object\\
22: observed user: Name of the owner of the object (or ``unknown'' if unknown)\\
23:\\
24: Please also consider the following information on other objects.\\
25: Use only other information related to the object being presented.\\
26:\\
27: other object:[\textcolor{teal}{\textbf{OTHER\_OBJECT\_LIST}}]\\
28:\\
29: Do not output anything other than the question. Keep your question brief.\\
30: Do not include coordinate information in the question text.\\
31:\\
32: object in question:[\textcolor{teal}{\textbf{TARGET\_OBJECT\_LIST}}]
\tcblower
\textcolor{teal}{\textbf{OTHER\_OBJECT\_LIST}}: List of Objects in the Environment\\
(Class, Attribute, Point.X, Point.Y, Observed User)\\
\textcolor{teal}{\textbf{TARGET\_OBJECT\_LIST}}: List of Objects Targeted for Questioning\\
(Class, Attribute, Point.X, Point.Y, Observed User)\\
\\
Example of \textcolor{teal}{\textbf{OTHER\_OBJECT\_LIST}} and \textcolor{teal}{\textbf{TARGET\_OBJECT\_LIST}}\\
-Refrigerator, [0,0,0,0,0,0,0,0,0,0,0,0,0,0,0,0,1,0,0,0,0,0,0,1,1,0,0,0,1,0], \\
0.736902236, 3.799175403, unknown \\
-Backpack, [1,0,0,0,0,0,0,0,0,0,0,0,0,0,0,0,0,0,0,0,0,0,1,0,0,1,0,0,1,0], \\
6.325937769, 2.471300272, hashimoto\\
-Book, [0,0,1,0,0,0,0,0,0,0,0,0,0,0,0,0,0,0,1,0,0,0,0,0,0,0,1,0,1,0], \\
6.517469798, 2.656295813, unknown
\end{prompt}

Additionally, when interpreting user answers, the name of the respondent is included in the prompt. The prompt used for this process is shown in Prompt.~\ref{prompt:understand_user_prompt}, which enables the model to correctly handle both possessive expressions (e.g., ``It's mine'') and explicit references to individuals (e.g., ``It's Taro's'').

\begin{prompt}{Ownership Identification}{understand_user_prompt}
1: You are an excellent household robot.\\
2: Ask the following question about the owner of an object.\\
3:\\
4: Question text : ``\textcolor{teal}{\textbf{QUESTION\_TEXT}}''\\
5: To this question, the user gives the following answer.\\
6: User answer : ``\textcolor{teal}{\textbf{USER\_ANSWER}}''\\
7:\\
8: Identify the name of the owner of the object from this content.\\
9: However, if the user's answer indicates that the object is shared by multiple people, rather than a specific person, please output ``Shared''.\\
10:\\
11: The list of users is as follows:\\
12: user\_list = \textcolor{teal}{\textbf{USER\_LIST}}.\\
13:\\
14: However, the name of the user who is responding is shown below.\\
15: Responding user : ``\textcolor{teal}{\textbf{RESPONDING\_USER}}''\\
16:\\
17: Extract the answer according to the user's input and output the owner's name as follows.\\
18:\\
19: Example\\
20: answer\_output = ``hashimoto''\\
21:\\
22: User answers are subject to typos, in which case please choose the closest match.\\
23:\\
24: Do not output anything other than ``answer\_output''.\\
25: Let's consider them in this order.
\tcblower
\textcolor{teal}{\textbf{QUESTION\_TEXT}}: The question asked by the robot.\\
\textcolor{teal}{\textbf{USER\_ANSWER}}: The natural language answer provided by the user.\\
\textcolor{teal}{\textbf{USER\_LIST}}: The list of owners in the environment. \\
\textcolor{teal}{\textbf{RESPONDING\_USER}}: The owner's name as received from the user's answer, including the shared label ``Shared''.
\end{prompt}

\section{Experiment~1: Ownership Learning under Simplified Simulation Conditions}\label{exp1}
\subsection{Objective}
The objective of this experiment is to evaluate whether the proposed method can efficiently learn object ownership in a simple simulated environment. In addition, we conduct an ablation study to assess the contribution of each feature—such as object position and attributes—to the accuracy of ownership estimation.

\subsection{Conditions}\label{exp1_condition}
\begin{figure}[tb]
    \centering
    \includegraphics[width=0.9\linewidth]{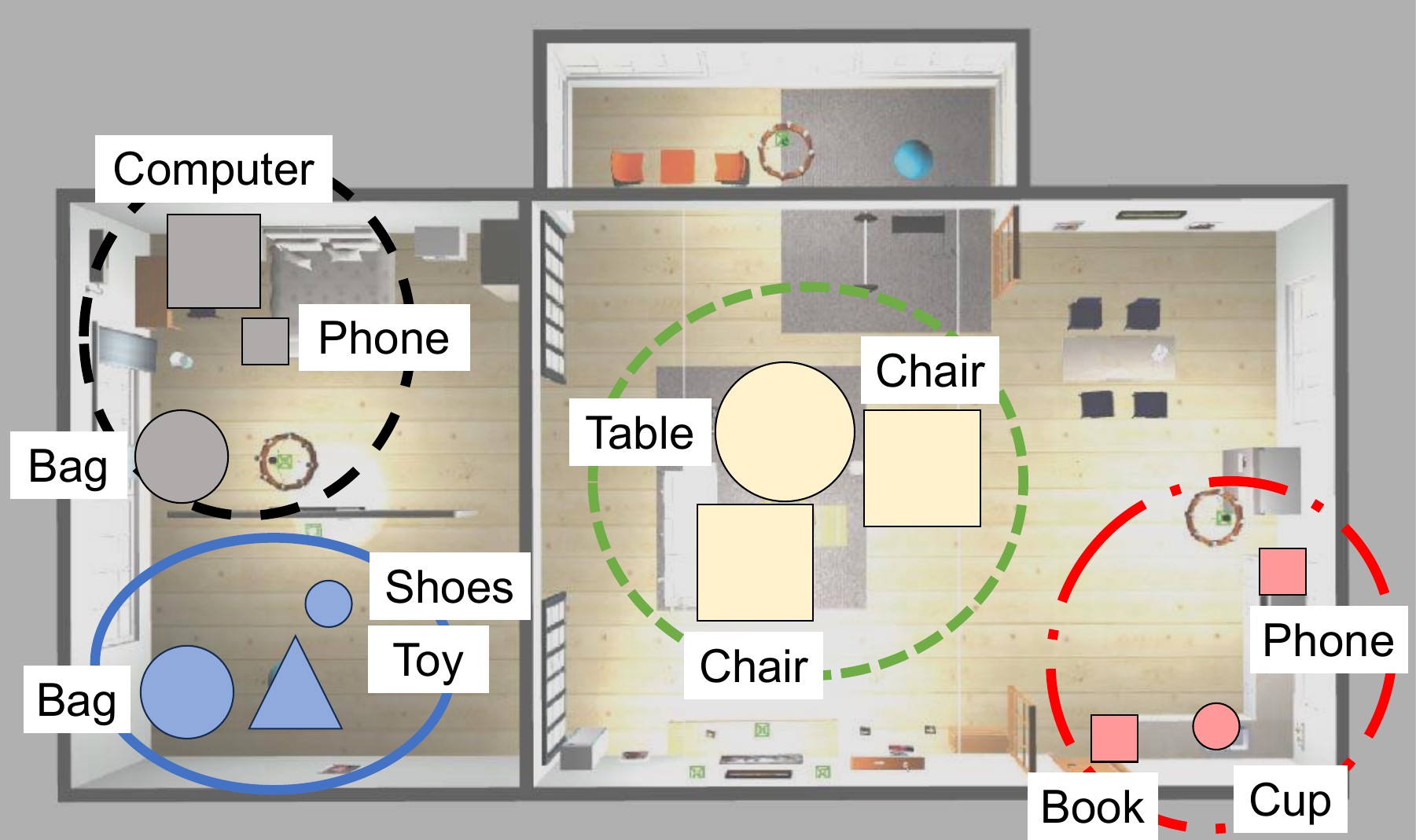}
    \caption{Object layout (Experiment~1): 
    Class name, color, size, shape, and position distribution of objects are shown}
    \label{map_1}
\end{figure}

The experiments were conducted in a Gazebo simulation environment using the aws-robomaker-small-house-world~\footnote{https://github.com/aws-robotics/aws-robomaker-small-house-world}.
We set up an environment with three users and 12 objects, each assigned an ownership label. 
The robot was assumed to have prior knowledge of all objects' 2D coordinates and category labels. 
The object layout is shown in Fig.~\ref{map_1}. 
Shared objects were labeled as ``Shared,'' while objects belonging to the same user were given consistent color attributes and placed in close proximity. 
To simulate realistic conditions, the environment included both owned and shared objects, as well as categories with multiple instances and those with a single instance.

Object positions were represented in a local two-dimensional coordinate system $(x, y)$. Each object was described by a 21-dimensional attribute vector constructed from four types of features:
\begin{itemize}
\item 
Class: one-hot vector corresponding to the number of object types (9 classes in this experiment).
\item
Color: 6-dimensional vector (red, blue, yellow, green, black, white).
\item 
Size: 3-dimensional vector (large, medium, small).
\item 
Shape: 3-dimensional vector (circle, square, triangle).
\end{itemize}
Attributes were manually assigned to each object.
User answers were collected in textual form.
This included explicit owner names (e.g., ``Taro's''), possessive pronouns (e.g., ``mine''), and referential expressions (e.g., ``my father's'').

During the learning process, the number of particles was set to 100, and the number of pseudo-observations (i.e., answer samples per question) was set to 10.
The hyperparameters were set as follows:
Attribute distribution parameter: $\alpha = 1.0$, 
User answer distribution parameter: $\beta = 0.01$, 
Ownership concept prior: $\gamma = 5.0$, 
Precision parameter for the position distribution: $\kappa_0 = 1.0$, 
Prior mean vector for the position distribution: $m_0 = [0, 0, 0, 0]$, 
Prior covariance matrix: $V_0 = \mathrm{diag}(0.1, 0.1)$
Degrees of freedom for the position distribution: $\nu_0 = 5.0$.
We also set the total number of ownership concepts: $L = 4$ and the total number of position distributions: $K = 4$
To stabilize estimation, the true position distribution index corresponding to each observed coordinate $x_n$ was fixed.

When integrating multimodal information—object position, attributes, and user answers—we treated user answers as the most reliable modality. 
A weighting coefficient $\omega_\text{answer} = 5.0$ was applied to user answers, while other modalities were equally weighted. 
For the LLM, we used GPT-4 (gpt-4-0613)~\cite{GPT-4}.

\subsection{Comparison Methods}
To evaluate the performance of our approach (ActOwL), we compare it with the following baseline methods:
\begin{enumerate}
    \item \textbf{IG-min}:
     A method that selects the object with the lowest IG for question generation and learns ownership knowledge based on the user's answer.
    \item \textbf{Random}:
    A method that randomly selects an object for question generation and learns ownership knowledge based on the user's answer.
    \item \textbf{No-LLM}:
    A method that selects objects for question generation based solely on the probabilistic generative model and IG, without using LLM, and learns ownership knowledge from the user's answers.
    \item \textbf{LLM-only}:
    A method that predicts ownership knowledge by prompting LLM without using any probabilistic generative model.
\end{enumerate}

In addition, to examine the contribution of each feature to ownership learning, we conducted the following ablation studies:
\begin{enumerate}
    \item 
    Only color information from the object's attributes is used.
    \item 
    Only the object's positional coordinates are used for learning, without incorporating attribute information.
    \item 
    Only the object's attribute information is used for learning, without incorporating positional coordinates.
\end{enumerate}

\subsection{Evaluation Metrics}
This study evaluates (1) the accuracy of ownership learning and (2) the efficiency of learning in terms of the number of user questions required. The latter focuses on how active exploration based on IG maximization improves learning efficiency in the probabilistic generative model.

\subsubsection{Adjusted Rand Index}
Ownership clustering accuracy was measured using the Adjusted Rand Index~(ARI), a standard metric for clustering performance.
ARI quantifies the agreement between predicted ownership clusters and ground-truth labels, with higher values indicating more accurate ownership concepts.

\subsubsection{Number of Questions}
Learning efficiency was evaluated by the number of questions posed by the robot to the user until a given ARI was achieved. This measure directly indicates how effectively the proposed method acquires ownership knowledge compared to baseline methods.

\subsection{Results}
\begin{figure}[tb]
    \centering
    \includegraphics[width=1.0\linewidth]{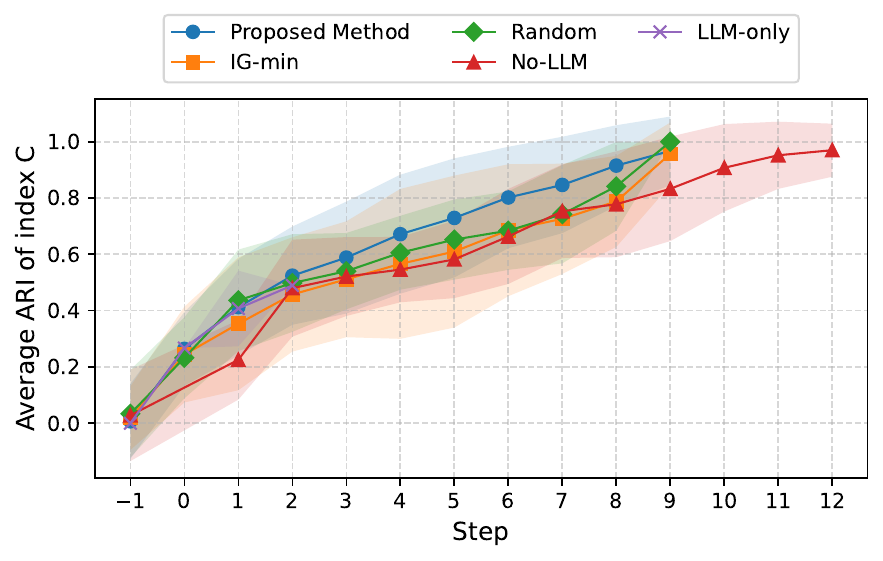}
    \caption{ARI of $C_n$ per step (Experiment~1): 
    Step $-1$ shows the post-exploration baseline, and Step 0 shows results after LLM-based shared/owned classification. Subsequent steps indicate performance after each additional question. Since shared objects are excluded from querying (except in No-LLM), learning typically converges by Step 9
    }
    \label{exp1_ari}
\end{figure}

Fig.~\ref{exp1_ari} shows ARI and its standard deviation for $C_n$ across 20 trials at each step. 
The proposed method consistently achieved the highest ARI values, indicating that the robot accurately learned ownership knowledge at an early step. 
This improvement stems from IG-based selection of informative objects, which enabled efficient acquisition of ownership knowledge. 
The No-LLM method also improved steadily but required more questions to reach similar accuracy, highlighting the benefit of excluding shared objects in advance. 
In contrast, the LLM-only method showed limited ARI improvement, despite asking fewer questions, suggesting that reliance on commonsense reasoning alone can lead to overgeneralization misaligned with actual ownership. 
By integrating LLM-based commonsense knowledge with probabilistic generative modeling, the proposed method achieved more robust and efficient ownership learning.

\begin{figure}[tb]
    \centering
    \includegraphics[width=1.0\linewidth]{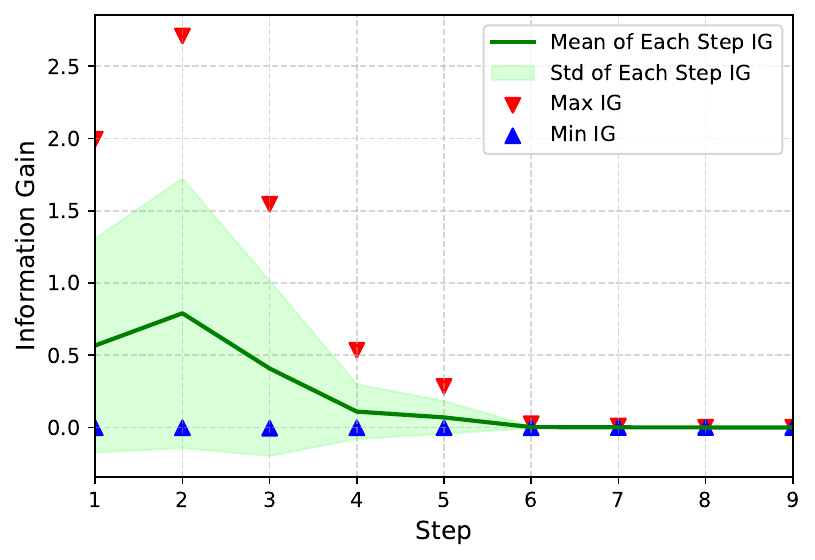}
    \caption{Trends in IG values per step (Experiment~1)}
    \label{exp1_ig}
\end{figure}

Fig.~\ref{exp1_ig} shows the transition of IG values for selected objects at each step across 20 trials. 
In the early steps, objects with high IG values were consistently chosen, demonstrating that question generation was strategic and efficient.
From Step 6 onward, IG variation decreased, suggesting that the remaining candidates contributed less new information. 
This indicates that the robot had already acquired sufficient knowledge by this point and could maintain high ownership prediction performance without additional questions.

\begin{figure}[tb]
    \centering
    \includegraphics[width=1.0\linewidth]{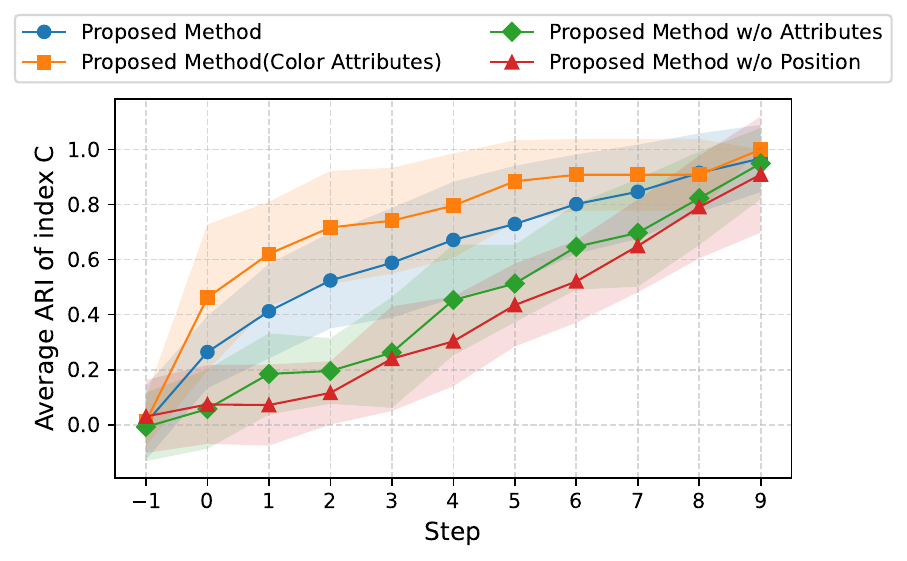}
    \caption{ARI of $C_n$ per step (Experiment~1: ablation study)}
    \label{exp1_ablation_ari}
\end{figure}

Fig.~\ref{exp1_ablation_ari} shows the average ARI for $C_n$ at each step across 20 trials in the ablation experiments, with error bars indicating standard deviation. 
Among the ablation settings, using only color information yielded the highest ARI, reflecting consistent color preferences among owners in the simulated environment. 
However, since real-world owners do not necessarily use objects of uniform color, relying solely on color lacks generalizability. 
In contrast, the proposed method, which integrates multiple attribute types, shows greater robustness and adaptability to diverse environments. 
Settings that used only positional information or only attribute information produced lower ARI than the proposed method, indicating that a single modality is insufficient and that combining positional and attribute information is essential for accurate ownership inference.

\section{Experiment~2: Ownership Learning in a Complex Simulated Environment}\label{exp2}
\subsection{Objective}
The objective of this experiment is to evaluate the effectiveness of the proposed method in a more complex environment. 
In particular, we examine whether it remains effective for ownership learning when both the number of objects and users are increased, simulating a more realistic household setting.

\subsection{Conditions}
\begin{figure}[tb]
    \centering
    \includegraphics[width=1.0\linewidth]{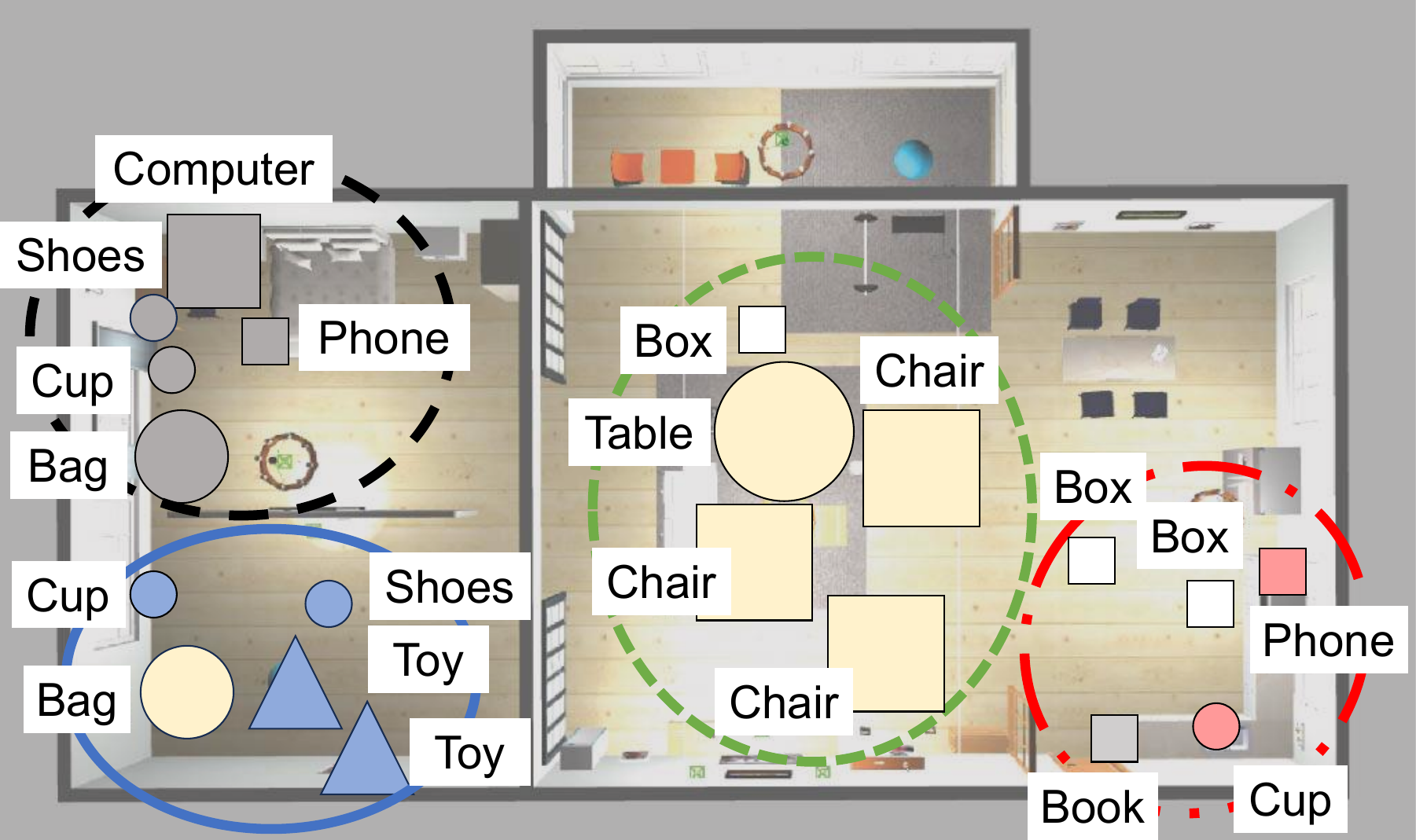}
    \caption{Object layout (Experiment~2): 
    Class name, color, size, shape, and position distribution of objects are shown}
    \label{map_2}
\end{figure}

The experiment was conducted in the same Gazebo simulation environment as Experiment~1, using the aws-robomaker-small-house-world. 
We set up an environment with three users and 20 objects, each assigned an ownership label. 
The robot was assumed to have prior knowledge of all objects' 2D coordinates and category labels through prior exploration. 
The object layout is shown in Fig.~\ref{map_2}. As in Experiment~1, the environment included both owned and shared objects, with categories appearing multiple times as well as only once. 
To evaluate robustness, most objects owned by the same user were grouped by consistent color attributes, while a subset was intentionally assigned inconsistent color patterns to introduce contradictory cues for ownership.

In this experiment, the number of object categories was set to 10, resulting in a 22-dimensional attribute vector constructed in the same manner as in Experiment~1. The representation of object positions and the format of user answers were also the same as in Experiment~1. 
All other experimental settings, including hyperparameters, multimodal integration, and LLM-based processing, were identical to those described in Section~\ref{exp1_condition}.

To evaluate performance, we employed the same comparison methods as in Experiment~1. 
The same evaluation metrics were also used to assess ownership learning accuracy and learning efficiency.

\subsection{Results}
\begin{figure}[tb]
    \centering
    \includegraphics[width=1.0\linewidth]{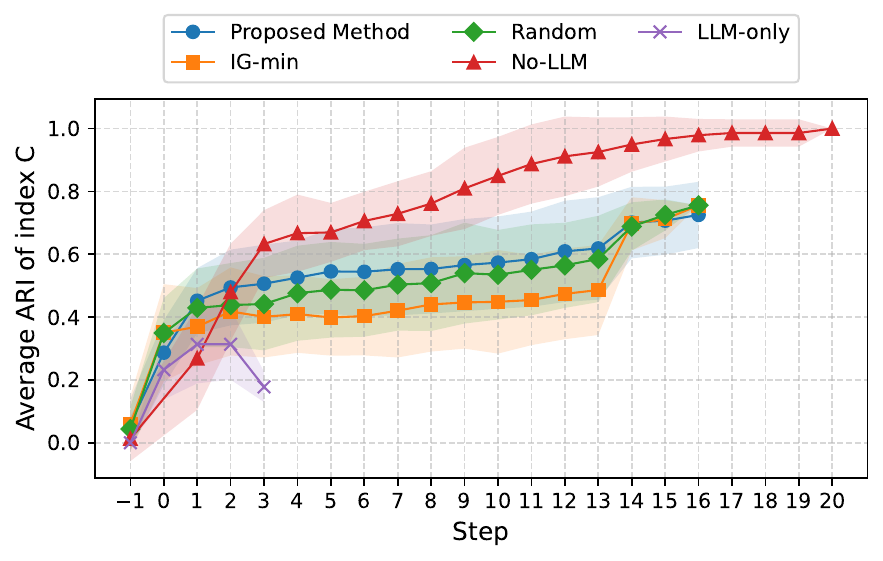}
    \caption{ARI of $C_n$ per step (Experiment~2): 
    Step $-1$ shows the post-exploration baseline, and Step 0 shows results after LLM-based shared/owned classification. Subsequent steps indicate performance after each additional question. Since shared objects are excluded from querying (except in No-LLM), learning typically converges by Step 16
    }
    \label{exp2_ari}
\end{figure}

Fig.~\ref{exp2_ari} shows the average ARI and standard deviation of $C_n$ over 20 trials at each step. 
In this experiment, the proposed method underperformed compared to No-LLM. 
This was mainly because the object class “Box,” although owned, was incorrectly categorized as shared and excluded from questioning, which resulted in missing information.
This reveals a limitation of the proposed strategy: reliance on initial classification can cause critical information loss and suppress learning performance. 
By contrast, No-LLM, which relies solely on the probabilistic generative model, achieved high ARI, indicating the robustness of the model even when positional and attribute consistency across owners is limited. 
Nevertheless, the proposed method outperformed the other baselines such as Random and IG-min, confirming that IG-based question selection contributes to efficient ownership learning.

\section{Experiment~3: Effectiveness in a Complex Real Environment}\label{exp3}
\subsection{Objective}
The objective of this experiment is to evaluate whether the proposed method can efficiently learn object ownership in a complex real-world environment. 
In particular, laboratory settings are often shared by multiple users with similar workspaces, and object attributes are not always clearly distinguishable. 
This experiment examines whether the method can still acquire ownership knowledge effectively under such challenging conditions.

\subsection{Conditions}
\begin{figure}[tb]
    \centering
    \includegraphics[width=1.0\linewidth]{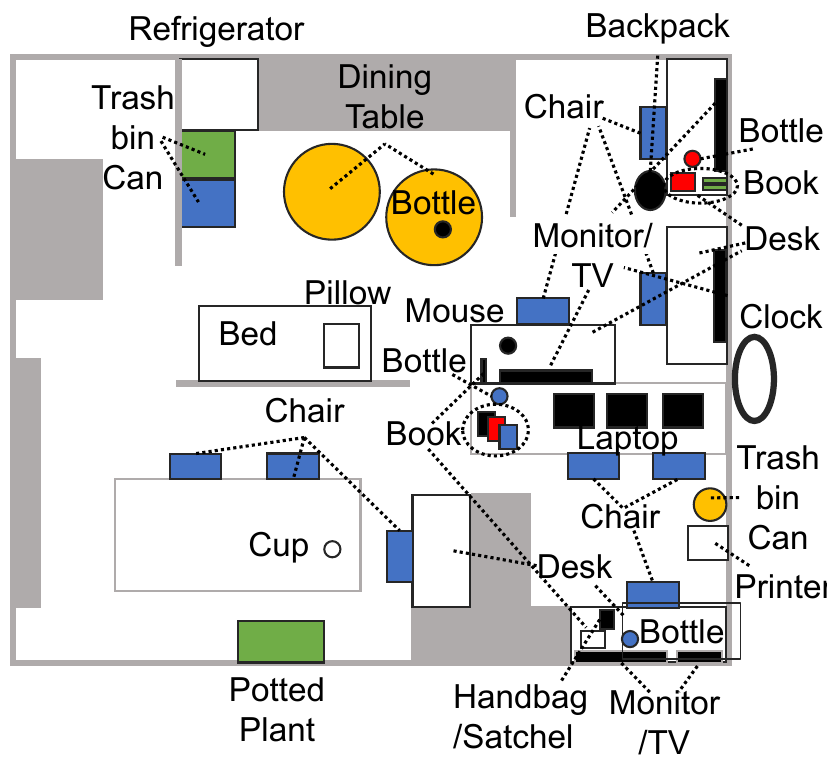}
    \caption{Object layout (Experiment~3): 
    Class name, color, size and shape of objects are shown}
    \label{map_3}
\end{figure}

\begin{figure}[tb]
    \centering
    \includegraphics[width=0.8\linewidth]{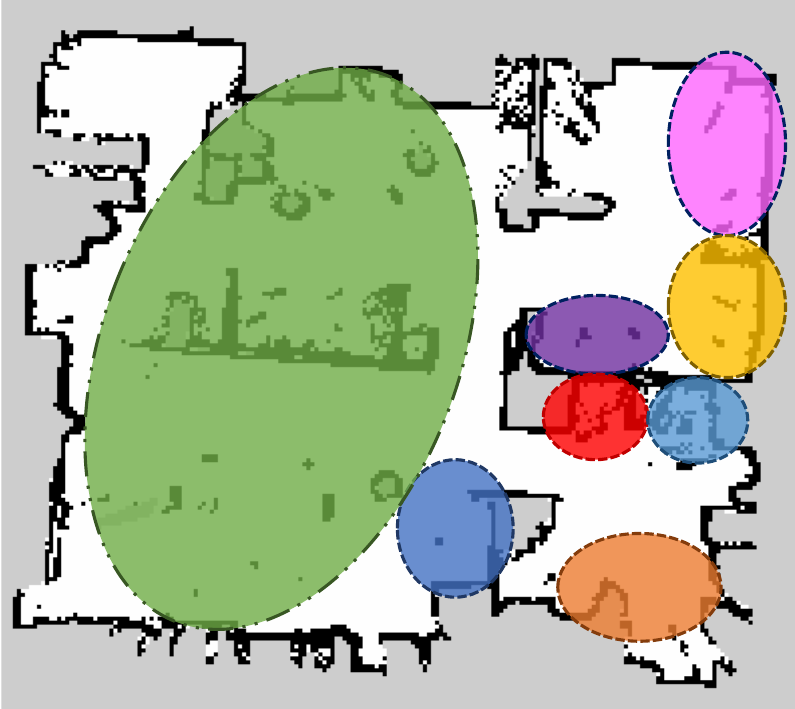}
    \caption{Position distribution (Experiment~3): 
    Large green circles on left show the distribution of shared objects, whereas the others show the distributions of owned objects
    }
    \label{map_3_position_distribution}
\end{figure}

The experiment was conducted in a real laboratory at Ritsumeikan University using the Human Support Robot (HSR)~\cite{HSR}. 
We set up a scenario with seven users and 48 objects, each assigned an ownership label. 
The robot was assumed to have prior knowledge of all objects' 2D coordinates and category labels through prior exploration. 
The object layout is shown in Fig.~\ref{map_3}, and the corresponding position distribution in Fig.~\ref{map_3_position_distribution}.

In this real-world environment, it is common for multiple users to share the same workspace, and objects used by different individuals often lack distinguishing visual features. 
As a result, object attributes are less effective cues for ownership, making spatial location and user answers more critical. 
To ensure realistic evaluation, we used actual laboratory objects with known ownership assignments.

The number of object categories was set to 18, resulting in a 30-dimensional attribute vector constructed in the same manner as described in Section~\ref{exp1_condition} for Experiment~1. Object positions and user answers were handled in the same way as in Experiment~1.

Most hyperparameter settings were identical to those in Experiment~1, except that the total number of ownership concepts was set to $L = 8$ and the total number of position distributions to $K = 8$. 
As before, the index $i_n$ of the spatial distribution corresponding to each observed coordinate $x_n$ was fixed for stability, and multimodal integration applied a weight of $\omega_\text{answer} = 5.0$ to user answers.
Furthermore, since no significant differences were observed in object attributes in the experimental environment, the attribute modality was down-weighted with a coefficient of $\omega_\text{attribute} = 0.1$, while the other modalities were equally weighted.

For LLM-based processing, we used GPT-4 (gpt-4-0613), as in the previous experiments. 
The prompt used for classifying objects as shared or owned (Prompt.~\ref{prompt:classify_object_prompt}) was slightly modified to reflect the contextual characteristics of the laboratory environment, as shown in Prompt.~\ref{prompt:prompt_for_exp3}.

\begin{prompt}{Prompt added in Experiment~3. }{prompt_for_exp3}
In Prompt.~\ref{prompt:classify_object_prompt}\\
8: However, the environment shall be a university laboratory.\\
9: Each student has his or her own desk in one room, and there is a shared space for meetings.
\end{prompt}

To evaluate performance, we employed the same comparison methods as in Experiment~1. 
The same evaluation metrics were also used to assess ownership learning accuracy and learning efficiency.

\subsection{Preparation before experiment}
In this experiment, the robot first constructed a map of the domestic environment using gmapping, a ROS-based mapping package. 
For object detection, it employed Detic~\cite{Detic} to recognize objects and obtain their category labels.

\subsection{Results}
\subsubsection{Qualitative evaluation}
Table~\ref{exp3_shared_owned} presents examples of object classes classified as shared or owned by the LLM. 
Incorporating the laboratory context into the prompt enabled appropriate classification based on usage and placement, showing that the LLM's commonsense knowledge functioned effectively. 
By pre-classifying shared objects, the robot excluded them from question generation, avoiding unnecessary queries. 
Consequently, ownership knowledge was learned more efficiently with fewer user questions.

\begin{table}[tb]
\centering
\caption{Examples of object classes classified as Shared and Owned objects (Experiment~3)}
\label{exp3_shared_owned}
\begin{tabular}{|c|p{5cm}|}
\hline
\textbf{Object Type} & \textbf{Examples} \\ \hline
Shared & Clock, Dining Table, Printer, Potted Plant, Refrigerator, Trash bin Can \\ \hline
Owned & Backpack, Bed, Book, Bottle, Chair, Cup, Desk, Handbag/Satchel, Laptop, Monitor/TV, Mouse, Pillow \\ \hline
\end{tabular}
\end{table}

\begin{figure}[tb]
    \centering
    \includegraphics[width=1.0\linewidth]{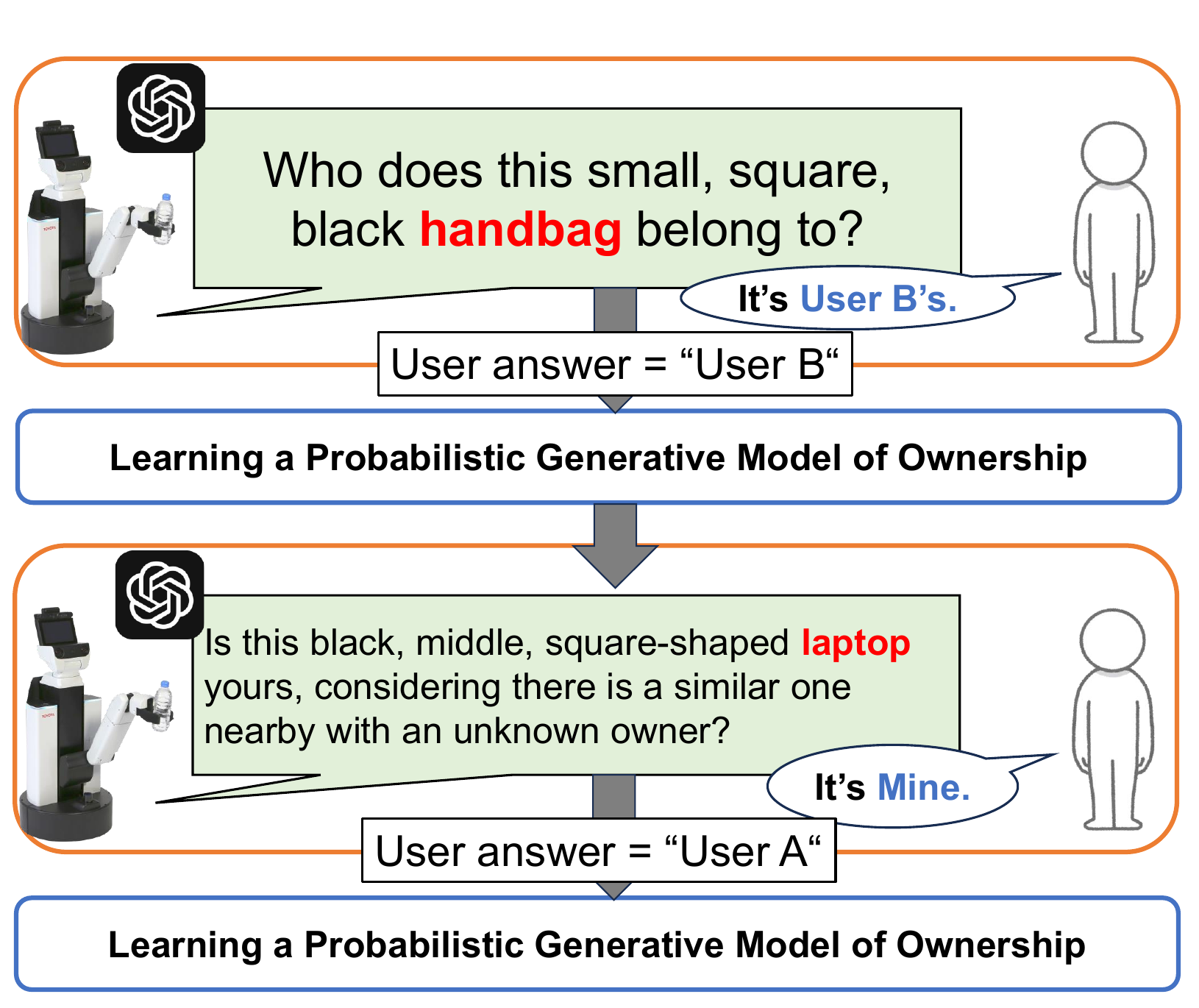}
    \caption{Sample questions generated by LLM}
    \label{question_generation}
\end{figure}

Fig.~\ref{question_generation} presents examples of questions generated in this experiment. 
The proposed method produced clear and flexible questions by leveraging spatial relationships with nearby objects. 
However, in some cases, objects not actually close were incorrectly described as ``nearby,'' indicating that the LLM could not fully capture spatial relationships. 
This highlights a remaining challenge in precise spatial reasoning.

\subsubsection{Quantitative evaluation}
\begin{figure}[tb]
    \centering
    \includegraphics[width=1.0\linewidth]{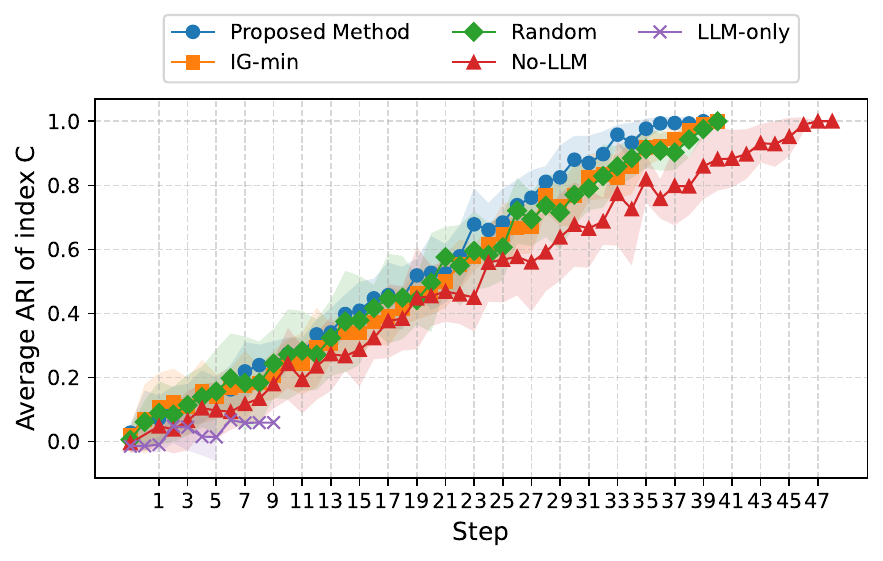}
    \caption{ARI of $C_n$ per step (Experiment~3):
    Step $-1$ shows the post-exploration baseline, and Step 0 shows results after LLM-based shared/owned classification. Subsequent steps indicate performance after each additional question. Since shared objects are excluded from querying (except in No-LLM), learning typically converges by Step 40
    }
    \label{exp3_ari}
\end{figure}

Fig.~\ref{exp3_ari} shows the average ARI and standard deviation of $C_n$ over 10 trials at each step. 
In the laboratory environment, where object types and appearances were highly similar across users, the discriminative power of attribute information was limited, making IG-based question selection less effective. 
Even so, by lowering the weight of the attribute modality, our method achieved slightly higher ARI than IG-min and Random, suggesting that adjusting modality contributions can improve adaptability in challenging environments.
Moreover, the proposed method outperformed both No-LLM and LLM-only in clustering consistency, indicating that the combination of strategic question generation and the probabilistic generative model remained effective. 
It also achieved accuracy comparable to or higher than No-LLM with fewer questions, showing that pre-classifying objects into shared and individually owned categories enhances learning efficiency beyond conventional approaches.
Overall, these results demonstrate both the potential of our approach and the inherent challenges of ownership learning in complex real-world settings, where users often possess visually similar objects and ownership boundaries are ambiguous.

\section{Limitations}\label{limitaion}
The method employed commonsense-based classification of objects into shared and owned categories using LLM; however, such interpretations can vary across environments, cultures, and contexts, leaving a risk of misclassification.
A more flexible classification framework is needed that accounts for uncertainty in sharedness.

The system assumed that user responses were always accurate and consistent; however, it was unable to handle ambiguous expressions or subjective variability.
To ensure robust dialogue in real-world scenarios, response processing must incorporate uncertainty in user utterances.

Object location, class, and attribute information were entirely annotated by hand, and the robot was not able to autonomously explore or perceive the environment.
In addition, feature weights were manually set; therefore, a mechanism is necessary that allows the robot to autonomously select and utilize relevant features based on context.

Ownership was treated as a binary relation—either individually owned or shared by all—and partial or group ownership could not be represented.
To reflect realistic social contexts, a more flexible framework is needed for representing and inferring complex ownership relationships.

\section{Conclusion}\label{conclusion}
\subsection{Summary}
In this study, we proposed ActOwL, which enables robots to efficiently learn object ownership in domestic environments. 
The approach selects objects based on IG and generates targeted questions to acquire ownership knowledge. 
By leveraging commonsense reasoning from an LLM, the robot can distinguish shared from owned objects, reducing the scope of user interaction. 
We evaluated ActOwL in both simulated and real-world settings, where it outperformed baseline approaches by achieving higher ownership clustering accuracy with fewer interactions. These results suggest that efficient acquisition of ownership knowledge through question generation can enhance a robot's ability to perform flexible and socially appropriate tasks in daily life environments.

\subsection{Future Work}
By leveraging background information about users in the environment, such as their roles and occupations, it becomes possible to narrow down candidate objects that are likely to belong to them. 
This can improve the precision of question generation by the LLM and is expected to enhance the efficiency of ownership learning.

The future framework will incorporate temporary ownership or dynamic changes in ownership over time. 
By integrating observations of user behavior and activity patterns within the environment, robots will be able to update ownership knowledge more flexibly and adaptively.

Once a robot understands ownership knowledge, it can immediately identify the target object when the user gives a command, enabling prompt and appropriate actions. 
This capability can support personalized and socially appropriate assistance in everyday tasks such as tidying up and managing personal belongings.

\section*{Acknowledgments}
This work was supported by JSPS KAKENHI Grants-in-Aid for Scientific Research (Grant Numbers
JP23K16975, JP25K15292) and JST Moonshot Research \& Development Program (Grant Number
JPMJMS2011).



\bibliographystyle{spmpsci_change}
\bibliography{articles}

\end{document}